\documentclass{article}

\usepackage{PRIMEarxiv}

\usepackage[utf8]{inputenc} 
\usepackage[T1]{fontenc}    
\usepackage{hyperref}       
\usepackage{url}            
\usepackage{booktabs}       
\usepackage{amsfonts}       
\usepackage{nicefrac}       
\usepackage{microtype}      
\usepackage{lipsum}
\usepackage{fancyhdr}       
\usepackage{graphicx}       
\usepackage{enumitem} 
\usepackage{float}
\usepackage{amsmath}
\graphicspath{{media/}}     

\title{Struc2mapGAN: improving synthetic cryo-EM density maps with generative adversarial networks
}

\author{
  Chenwei Zhang \textsuperscript{1} \ \ \ \ \ \ \ \ \ \
  Anne Condon \textsuperscript{1} \ \ \ \ \ \ \ \ \ \
  Khanh Dao Duc \textsuperscript{2} \\
  \textsuperscript{1} Department of Computer Science, UBC \;\;\;\;\;
  \textsuperscript{2} Department of Mathematics, UBC\\
  \texttt{\{cwzhang, condon\}@cs.ubc.ca} \;\;\;\;\;\;\;\;\;\;\;\;\;\;
  \texttt{kdd@math.ubc.ca}\\
}

\begin{document}
\maketitle


\begin{abstract}
Generating synthetic cryogenic electron microscopy 3D density maps from molecular structures has potential important applications in structural biology. Yet existing simulation-based methods cannot mimic all the complex features present in experimental maps, such as secondary structure elements. As an alternative, we propose \emph{struc2mapGAN}, a novel data-driven method that employs a generative adversarial network to produce improved experimental-like density maps from molecular structures. More specifically, \emph{struc2mapGAN} uses a nested U-Net architecture as the generator, with an additional L1 loss term and further processing of raw training experimental maps to enhance learning efficiency. While \emph{struc2mapGAN} can promptly generate maps after training, we demonstrate that it outperforms existing simulation-based methods for a wide array of tested maps and across various evaluation metrics.
\end{abstract}


\section{Introduction}
Cryogenic electron microscopy (cryo-EM) is a powerful technique for structural determination of biological macromolecules such as proteins and nucleic acids, which has been widely used in structure-based drug discovery \cite{dd1,dd2}. 
Cryo-EM produces a series of 2D images that are then reconstructed into electron density maps, providing a 3D voxelized representation of the macromolecular complex. 

For the past few years, deep learning-based methods have been introduced to automate the construction of macromolecular atomic models from density maps (i.e. 3D coordinate structure) \cite{modelangelo,deeptracer,embuild}. 
However, most of these methods require high-fidelity experimental maps, given  the lack of reliable experimentally derived map-model pairs at lower resolutions \cite{wriggers2015numerical}. In this context, generating accurate synthetic maps can help to include the vast majority of structures that are limited to be solved at lower resolutions.
In addition, these can also be used for ``rigid-body fitting'' in the building process (i.e. aligning a whole or partial atomic model into a density map \cite{empot,eman1}),
 ``sharpening'' density maps \cite{emready,cryofem} (i.e. enhancing the map to facilitate model building) \cite{embuild}, or providing more ground-truth training targets.
 
As generating synthetic cryo-EM density maps has become a pivot, various simulation-based methods are available \cite{eman1, eman2,chimerax,tempy2,situs}, essentially using a resolution-lowering point spread function to convolute atom points extracted from PDB structures (atomic models). Upon treating atoms independently, these methods may fail to characterize more complex features, such as secondary structure elements (SSEs), interatomic interactions, or specific image artifacts inherent to the 3D reconstruction process. 
In addition, Alshammari et al. \cite{alshammari2022refinement} proposed an approach to generate cryo-EM maps by applying a Gaussian convolution to an experimental map rather than deriving from an atomic structure, allowing for the capture of complex features.
As an alternative, we thus introduce a data-driven approach that leverages generative models. Our method, called \emph{\textbf{struc2mapGAN}}, uses a generative adversarial network (GAN) to enhance the generation of experimental-like cryo-EM density maps from PDB structures. Our contributions are as follows.
\begin{itemize}
    \item We introduce the first deep learning-based method, to our knowledge, for generative modeling of cryo-EM density maps. Our method enhances the learning efficiency by curating high-quality training data and alleviates mode collapse \cite{modecollapse} in GANs through the integration of \texttt{SmoothL1Loss} into the model. 
    \item Our benchmarking shows superior overall performance against simulation-based methods, across various evaluation metrics (e.g. correlation) and over a variety of tested maps. Our experiments also suggest that this performance improvement results from better capture of SSEs.
    \item We benchmark the runtime of \emph{struc2mapGAN} and demonstrate its practical suitability for generating large-scale maps.
\end{itemize}

\section{Related work}

\subsection{Existing simulation-based methods}

Simulation-based methods for generating density maps from their associated PDB structures are based on the convolution of atom points with resolution-lowering point spread functions such as Gaussian, triangular, or hard-sphere \cite{spread1, spread2}. 
Given a PDB structure containing $M$ atoms, a general Gaussian simulation formula for producing a density value $\rho$ at a grid point $\mathbf{x}$ is expressed as:
\begin{equation}
    \rho(\mathbf{x}) = \sum_{i=1}^{M} \theta Z_i e^{-k|\mathbf{x}-\mathbf{r}_i|^2},
\end{equation}
where $Z_i$ represents the atomic number and $\mathbf{r}_i$ is the position vector of the i-th heavy atom, $\theta$ is a scaling factor and $k$ is defined in terms of the resolution \cite{emready,embuild}. 
Such methods including \emph{e2pdb2mrc} in EMAN2 \cite{eman2} (originally called \emph{pdb2mrc} in EMAN1 \cite{eman1}), \emph{molmap} in UCSF ChimeraX \cite{chimerax}, \emph{StructureBlurrer} in TEMPy2 \cite{tempy2}, and \emph{pdb2vol} in Situs \cite{situs}, generate 3D density maps using a Gaussian point spread function, where the real-space dimension corresponds to the desired resolution, which varies based on the specific resolution convention of each package.

\subsection{Generative adversarial networks}
GAN is one of the most prevalent generative models widely employed in image generation, super-resolution, and 3D object generation \cite{gan}. The GAN architecture comprises two main models, a generator \emph{$G$} and a discriminator \emph{$D$} trained together. \emph{$G$} generates a batch of images (we call them fake images), and these images are fed along real images (ground-truth reference images) into \emph{$D$} to be classified as real or fake. During training, \emph{$G$} generates fake images to fool \emph{$D$}, while \emph{$D$} is updating its parameters to discriminate the fake ones. 
Mathematically speaking, the two models \emph{$G$} and \emph{$D$} compete in a two-player minimax game with the objective function $L(G,D)$: 
\begin{equation}
    \min_G \max_D V(D, G) = \mathbb{E}_{\mathbf{x}} [\log D(\mathbf{x})] + \mathbb{E}_{\mathbf{z}} [\log (1 - D(G(\mathbf{z})))].
\end{equation}
The generator minimizes the $\log (1 - D(G(\mathbf{z})))$ term predicted by \emph{$D$} for fake images. Conversely, the discriminator maximizes the log probability of real images, $\log D(\mathbf{x})$ and the log probability of correctly identifying fake images, $\log (1 - D(G(\mathbf{z})))$.
Over the past few years, the GAN architecture has been adapted for various purposes and has demonstrated superior performance in multiple domains \cite{wgan,cyclegan,cryogan,pix2pix,emgan}.

\section{Materials and Methods \label{methods}}

\begin{figure*}[ht]
    \centering
    \includegraphics[width=\linewidth, trim={0cm 1cm 0 0cm}, clip]{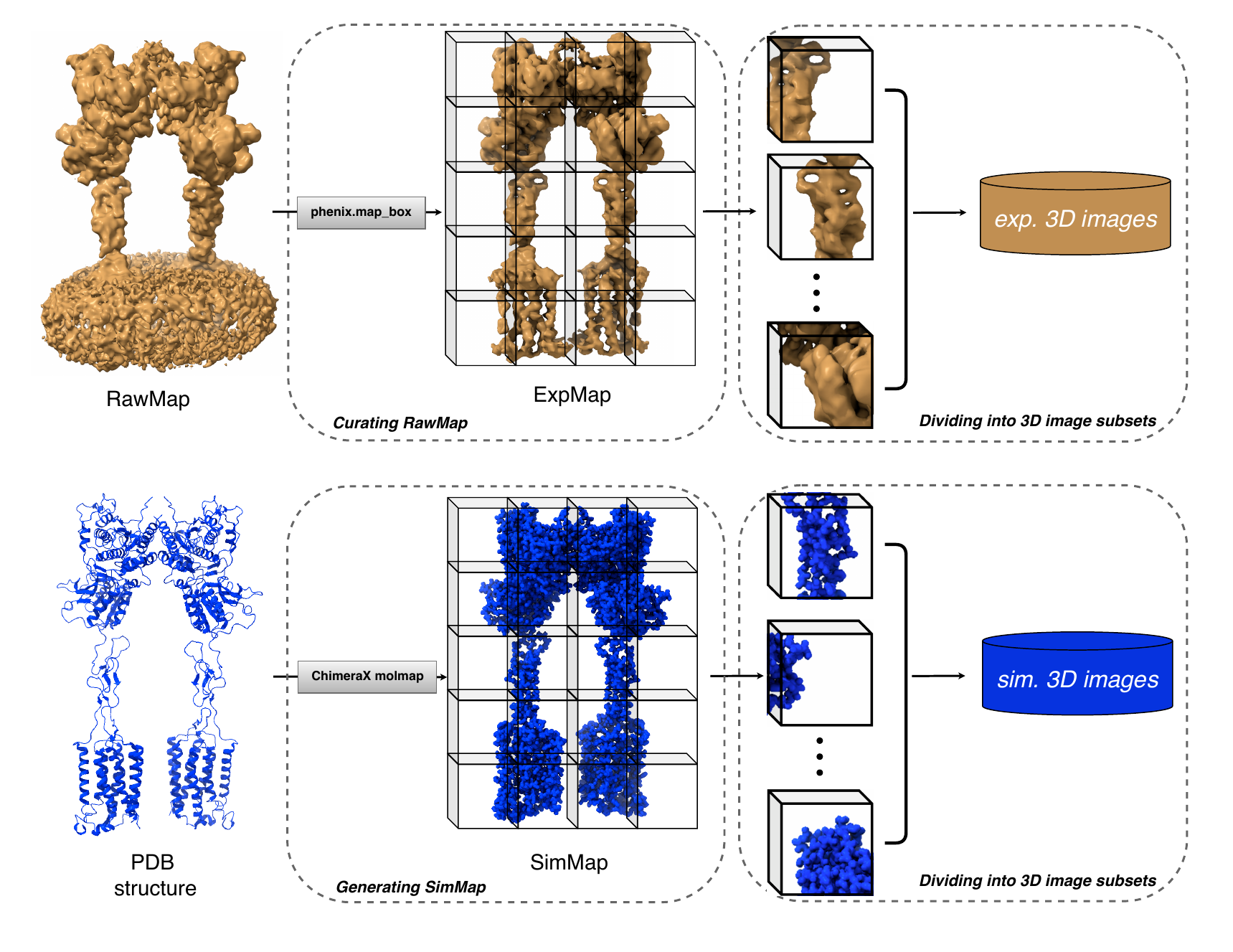}
    \caption{ 
    The data preprocessing workflow. The top panel depicts the process of curating raw experimental maps and dividing the curated maps into 3D image subsets as training targets. The bottom panel depicts the process of generating simulated maps and dividing them into 3D image subsets as training inputs. 
    }
    \label{fig:dataprocess}
\end{figure*}

\subsection{Datasets}

\paragraph{Training and validation data:}
To train and validate the GAN model, our dataset was built with a set of high-resolution cryo-EM density maps ranging from 2.17 {\AA} to 3.9 {\AA} from the EMDB databank \cite{EMDB} and associated PDB structures from the PDB databank \cite{PDB}. To ensure the density maps have proper alignments with their associated PDB structures, we removed maps from the dataset if:
(\romannumeral 1) maps contain extensive regions without corresponding PDB structures;
(\romannumeral 2) maps are misaligned with associated PDB structures;
(\romannumeral 3) maps contain various macromolecules, such as nucleic acids;
(\romannumeral 4) PDB structures only contain backbone atoms and/or include unknown residues.
In addition, pairs of simulated and experimental maps with correlation lower than 0.65 (calculated using ChimeraX) were excluded. 
To eliminate data redundancy, we measured the sequence identity between PDB structures and retained only one when the identity exceeded 30 \%.
After applying these filtering steps, a total of 149 cryo-EM density maps and associated PDB structures remained, as listed in Supplementary Table S1. 134 map-PDB pairs (90 \%) were selected as the training set and 15 pairs (10 \%) as the validation set. 

\paragraph{Test data:}
For the test set, we randomly selected 130 PDB structures from the PDB databank and associated cryo-EM density maps from the EMDB databank with resolution ranging from 3 {\AA} to 7.9 {\AA}, as detailed in Supplementary Table S2. Note that these 130 examples do not overlap with the training and validation data sets.

\subsection{Preprocessing:}
According to the following preprocessing steps illustrated in Figure \ref{fig:dataprocess}, both experimental maps (denoted as $Raw\-Maps$) and simulated maps from PDB structures (denoted as $Sim\-Maps$) were pre-processed to facilitate the GAN training. In addition, our GAN architecture also takes 3D images of reduced size as input.

\paragraph{Curating Raw\-Maps:}
The raw experimental maps consistently contain background noise, such as lipid solvents and artifacts from nanodiscs, which compromise the pairing accuracy between $Raw\-Maps$ and $Sim\-Maps$. To mitigate these effects, we applied a mask to the raw map to isolate the region containing only the protein structure. We first aligned the PDB structure with the corresponding map and employed \emph{phenix.map\_box} to create a rectangular box slightly larger than the targeted region. Subsequently, we resampled the masked map at a grid voxel size of $1 \text{\AA} \times 1 \text{\AA} \times 1 \text{\AA}$ and applied min-max normalization to scale the map's voxel values to the range of $[0, 1]$ to maintain uniformity. The curated map, termed $Exp\-Map$, was then employed for training the network. Note that this curation strategy to make $Exp\-Map$ targets significantly enhanced the performance of the GAN model, as evidenced in our ablation study (see section \ref{ablation}).

\paragraph{Generating Sim\-Maps:}
The input data $Sim\-Maps$ were simulated using the \emph{molmap} function. We converted the PDB structure into a simulated density map on a grid corresponding to the $Exp\-Map$, with a resolution cutoff at 2 {\AA}. This simulated map was then min-max normalized to the range of $[0, 1]$, aligning with its corresponding $Exp\-Map$. Moreover, in light of enhancing the model's robustness, we utilized \emph{TorchIO} \cite{torchio} to add random Gaussian noise, random anisotropy, and random blur to augment the input data.

\paragraph{Dividing maps into 3D image subsets:}
Considering the varying dimensions of each map and the constraints of GPU memory, we zero-padded the $Sim\-Maps$ and $Exp\-Maps$ and divided them into smaller 3D images (denoted as \textit{exp. 3D images} and \textit{sim. 3D images}, respectively) with dimensions of $32\times32\times32$.
To do so, we first created a padded map that exceeded the dimensions of the input map by $2 \times 32$ in each dimension, ensuring no boundary issues. The original map was then centrally placed within this padded map. Following this procedure yielded a total of 403710 3D images for training and 62635 for validation.

\subsection{The GAN architecture}
\begin{figure}[ht]
    \centering
    \includegraphics[width=0.9\linewidth, trim={0cm 10cm 12cm 0cm}, clip]{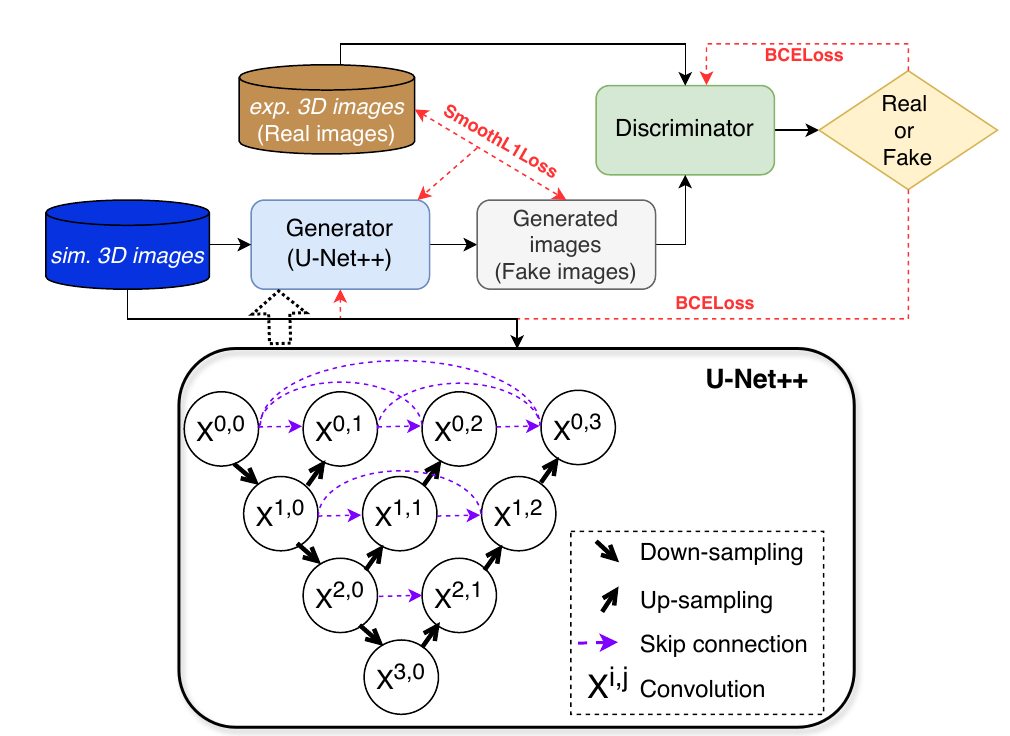}
    \caption{ 
    The \emph{struc2mapGAN} architecture. The bottom panel illustrates the U-Net++ architecture. $X^{i,j}$ refers to the convolution block at depth $i$ and position $j$ of the network.
     }
    \label{fig:architecture}
\end{figure}

Figure \ref{fig:architecture} depicts the \emph{struc2mapGAN} architecture that comprises a generator and a discriminator. 
We have implemented a nested U-Net architecture (U-Net++) \cite{unet++} as the generator (shown in the bottom panel of Figure \ref{fig:architecture}). The encoder and decoder blocks of the U-Net++ follow the same design, each consisting of 3D convolution layers with a kernel size of $3\times3\times3$. Unlike standard U-Nets, U-Net++ applies dense skip connections that effectively bridge the encoder and decoder feature maps, for enhanced gradient flow. 3D max-pooling layers, featuring a kernel size of 2 and a stride of 2, are employed for down-sampling, while trilinear interpolation layers with a scale factor of 2 are used for up-sampling.

The discriminator network has been designed for 3D volumetric data classification. It consists of four 3D convolution layers with a kernel size of $3\times3\times3$. An adaptive average-pooling layer is applied after the last convolution layer to reduce the feature map to the size of  $1\times1\times1$, which is then flattened and passed through three fully connected layers with ReLU activations between them. The final layer outputs a single value for binary classification.

For both the generator and discriminator, we employed a parametric rectified linear unit (PReLu) as the activation function between convolution layers, coupled with instance normalization. We opted for instance normalization over the more commonly used batch normalization as the statistics should not be averaged across instances within a batch. This consideration is crucial since local 3D image subsets of cryo-EM density may originate from different maps. The total number of parameters for the generator is 10.3 million and for the discriminator is 1.2 million.

\subsection{Training and inference of struc2mapGAN}

\paragraph{Training:}
During training, \emph{struc2mapGAN}accepts paired \textit{sim. 3D images} and \textit{exp. 3D images} as input and outputs modified images. 
To impose the generator to minimize the difference between the prediction and the target, we incorporated the standard adversarial loss for each input \textit{sim. 3D image} with the smooth L1 loss in the generator, as
\begin{equation}
    \text{SmoothL1Loss}(X,Y) =
    \begin{cases}
    0.5(X-Y)^2, & \text{if } |X-Y| < 1, \\
    |X - Y| - 0.5, & \text{otherwise},
    \end{cases}
\end{equation}
where $X$ is the GAN-predicted image, i.e. $G(\textit{sim. 3D image})$, and $Y$ is the paired target image, i.e. \textit{exp. 3D image}. The adversarial loss is calculated as the binary cross entropy loss of the discriminator's predictions, as
\begin{equation}
    \text{Loss}_{\text{adversarial}} = - \log (D(X)).
\end{equation}
Therefore, the generator loss is formulated as a linear combination of adversarial loss and smooth L1 loss:
\begin{equation}
    \text{Loss}_G = \text{SmoothL1Loss} + \alpha \times \text{Loss}_{\text{adversarial}},
\end{equation}
where $\alpha$ is a tuning hyperparameter. 
The discriminator loss adheres to the standard GAN loss formula, aiming to classify $X$ as fake and $Y$ as real:
\begin{equation}
    \text{Loss}_D = - (\log D(Y) + \log (1 - D(X))).
\end{equation}

\paragraph{Inference:}
To generate a map from a PDB structure using our trained generator, the input PDB structure was initially converted to a map following the procedure of generating $Sim\-Maps$.
This converted map was subsequently zero-padded and divided into 3D image subsets with dimensions of $32\times32\times32$, following the same strategy employed during training data preprocessing. These image subsets were then refined by the trained generator to yield post-processed images. The post-processed images were then reassembled back to their original map dimensions. To ensure that no spatial information was lost, we exclusively utilized the center $20\times20\times20$ voxels of each 3D image to reconstruct the map, in accordance with the method proposed by Si et al. \cite{cascadecnn}. 

\paragraph{Implementation:}
\emph{Struc2mapGAN} was implemented in PyTorch-2.2.2 + cuda-12.1, with all training and validating processes carried out on eight NVIDIA A100 GPUs, each with 80 GB VRAM. This setup supported a batch size of 128. The network was trained over $150$ epochs, with each epoch taking approximately $765$ seconds. The total training duration was approximately 1 day and 8 hours.
For optimization, NAdam \cite{nadam} optimizers were employed with a learning rate set at $0.0001$. 
We tested three different $\alpha$ values: $0.1, 0.01, 0.001$. Ultimately, $\alpha = 0.01$ was selected for yielding the best performance.

\subsection{Evaluation metrics}
To measure the accuracy of \emph{struc2mapGAN} in generating maps that are similar to experimental reference ones, we used the following evaluation metrics.

\paragraph{Structural similarity index measure:} 
The structural similarity index measure (SSIM) evaluates the similarity between two images based on three key features: luminance, contrast, and structure \cite{ssim}. This measure has been extended to evaluate 3D volumetric data and has been adapted as a loss function in cryo-EM density map studies \cite{emready,embuild}. The SSIM score between samples $X$ and $Y$ is calculated as follows:
\begin{equation}
    \text{SSIM}(X,Y) = \frac{(2\mu_{X}\mu_{Y}+c_1)(2\sigma_{XY}+c_2)}{(\mu_{X}^{2}+\mu_{Y}^{2}+c_1)(\sigma_{X}^{2}+\sigma_{Y}^{2}+c_2)},
\end{equation}
where $\mu_X$ and $\mu_Y$, $\sigma_X$ and $\sigma_Y$ are the means and variances of samples $X$ and $Y$, respectively. $\sigma_XY$ refers to the covariance of $X$ and $Y$. Constants $c_1$ and $c_2$ are included to stabilize the division. SSIM excels in capturing the local contrast and textures, which are crucial in cryo-EM density maps. In our study, we employed the \emph{scikit-image} package to calculate the SSIM between a \emph{struc2mapGAN}-generated or simulation-based density map and its corresponding experimental map. 

\paragraph{ChimeraX correlation:}
UCSF ChimeraX \cite{chimerax} provides a built-in \emph{measure correlation} function to calculate the correlation between two density maps in two ways. One way is to compute the correlation between two maps directly without considering the deviations from their means:
\begin{equation}
    \text{correlation}(X,Y) = \frac{\langle X, Y \rangle}{\|X\|\|Y\|},
\end{equation}
where $\langle X,Y\rangle$ represents the dot product of maps $X$ and $Y$; $\|X\|\|Y\|$ are the norms of the maps. The other way calculates the correlation between deviations of two maps from their means:
\begin{equation}
    \text{correlation about mean}(X,Y) = \frac{\langle X-X_{\text{ave}}, Y-Y_{\text{ave}} \rangle}{\|X-X_{\text{ave}}\|\|Y-Y_{\text{ave}}\|},
\end{equation}
where $X_{\text{ave}}$ and $Y_{\text{ave}}$ are the means of the two maps.
In the context of comparing cryo-EM density maps, we calculated the two correlations to assess the similarity across the entire volume of the two maps.

\paragraph{Pearson correlation coefficient:}
The Pearson correlation coefficient (PCC) is a measure of the linear correlation between two datasets \cite{pcc} and is defined by:
\begin{equation}
    \text{PCC}(X,Y) = \frac{\sum_{i=1}^{n} (X_i - \overline{X})(Y_i - \overline{Y})}{\sqrt{\sum_{i=1}^{n} (X_i - \overline{X})^2 \sum_{i=1}^{n} (Y_i - \overline{Y})^2}},
\end{equation}
where $X_i$ and $Y_i$ represent the individual data points in datasets $X$ and $Y$, respectively. $\overline{X}$ and $\overline{Y}$ are the means of the datasets, and $n$ is the total number of data points.  
In the context of comparing cryo-EM density maps, the PCC provides similarities between the two maps in terms of their density distributions.

\section{Results}

We conducted a comprehensive study to evaluate the performance of \emph{struc2mapGAN} using a test set of 130 PDB structures and associated experimental density maps (note that these structures and maps were not used for training the GAN network). In our benchmarking, simulation-based maps were generated using various methods (\emph{molmap}, \emph{e2pbd2mrc}, and \emph{StructureBlurrer}), with a resolution cutoff at 2 {\AA}.

\begin{figure}[ht]
    \centering
    \includegraphics[width=\linewidth, trim={1cm 11.5cm 2cm 1cm}, clip]{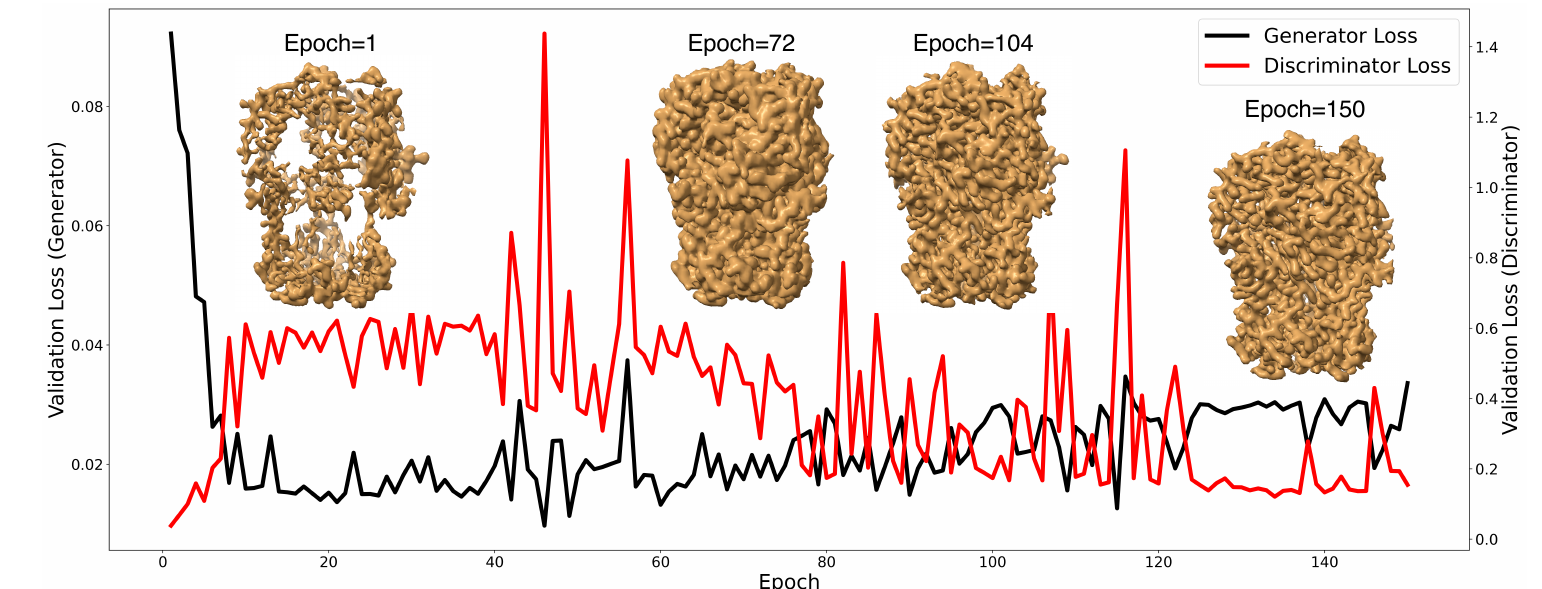}
    \caption{ 
    Validation loss curves of the generator and discriminator in black and red, respectively, with snapshots of generated maps from the models trained at specific epochs.
    }
    \label{fig:lossfig}
\end{figure}

\subsection{Learning curves with intermediate maps}
To analyze the training process, we visualized the training dynamics of \emph{struc2mapGAN} in Figure \ref{fig:lossfig}. The validation losses of both the generator (black) and discriminator (red) were plotted over 150 epochs. At the beginning of training, the generator loss decreased sharply, reflecting its quick adaptation to produce more realistic maps, while the discriminator loss slightly increased owing to the generator's enhancements. Throughout the training, fluctuations in both losses indicate continuous learning and adaptation. By the end of the training, the losses of both the generator and discriminator tended to converge, suggesting a balanced and stabilized adversarial training outcome. Alongside the loss curves, selected \emph{struc2mapGAN}-generated maps from epochs 1, 72, 104, and 150 were displayed. These intermediate maps visually encapsulated the gradual improvement in the quality and fidelity of the generated maps. For instance, the map from Epoch 1 showed incomplete densities. As training proceeded, the generated maps learned to represent complete densities, with their resolution progressively refined, as illustrated in the maps from Epoch 72 to Epoch 150.

\subsection{Visual comparison between generated maps by \emph{struc2mapGAN} and molmap}
\begin{figure}[!p]
    \centering
    \includegraphics[width=\linewidth, trim={0cm 3.5cm 0cm 0cm}, clip]{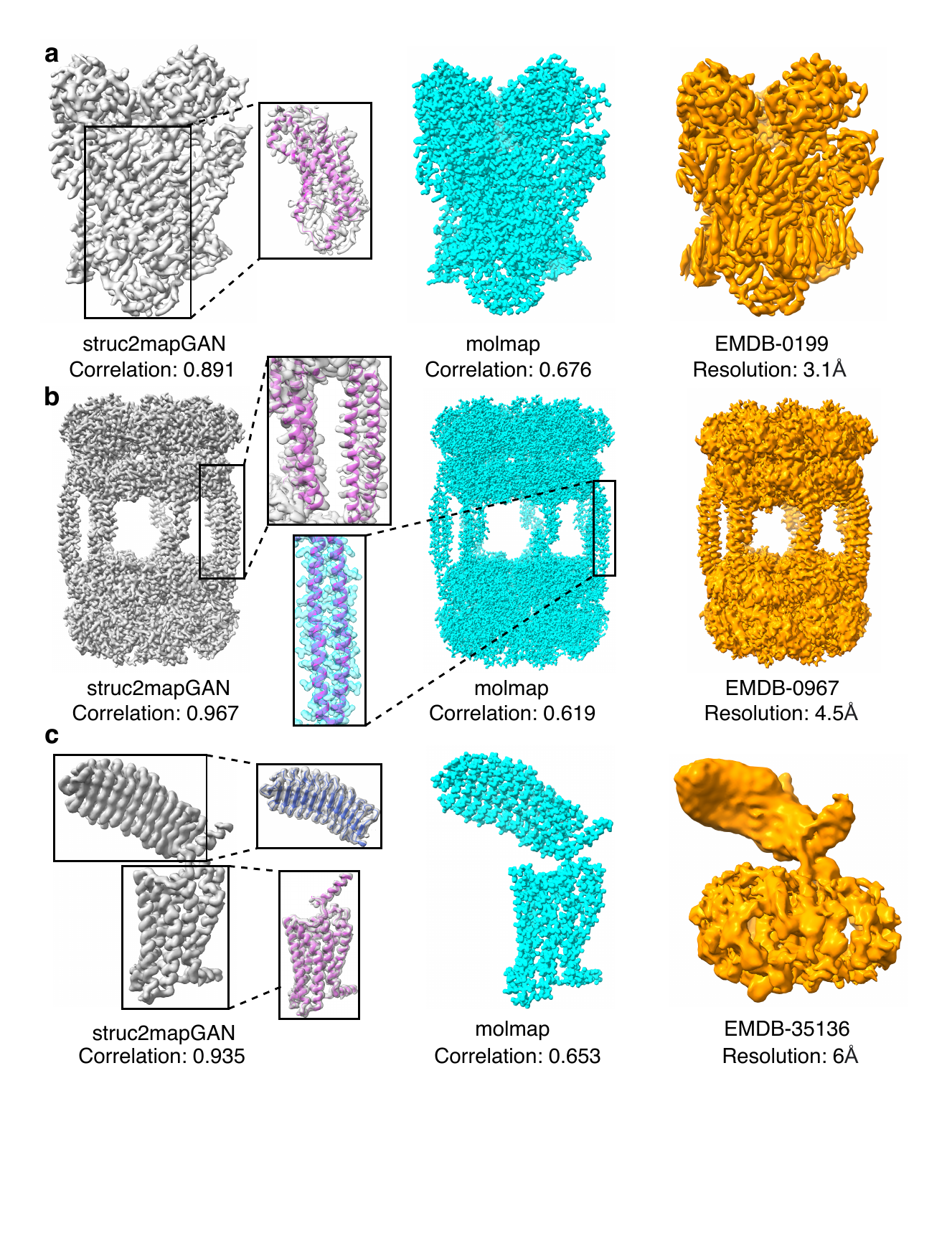}
    \caption{ 
    Examples of \emph{struc2mapGAN} (gray) and \emph{molmap} (cyan) generated maps, and the raw experimental maps (orange). The PDB structures of $\alpha$-helices (pink) and $\beta$-sheets (blue) are superimposed on the maps. 
    \textbf{a.} Human STEAP4 bound to NADP, FAD, heme and Fe(III)-NTA (EMDB ID: 0199; PDB ID: 6HCY; reported resolution: 3.1 {\AA}).
    \textbf{b.} AAA+ ATPase, ClpL from Streptococcus pneumoniae: ATPrS-bound (EMDB ID: 0967; PDB ID: 6LT4; reported resolution: 4.5 {\AA}).
    \textbf{c.} Follicle stimulating hormone receptor (EMDB ID: 35136; PDB ID: 8I2H; reported resolution: 6 {\AA}). Visualization of cryo-EM density maps and PDB structures was produced by UCSF ChimeraX \cite{chimerax}.
    }
    \label{fig:visual_comp}
\end{figure}

Upon training, we first visualized three GAN-generated density maps alongside their associated \emph{molmap}-generated maps, with corresponding experimental maps ranging from high to low resolutions. The visual comparisons are displayed in Figure \ref{fig:visual_comp}. 
We observed that as \emph{struc2mapGAN} is trained to represent high-resolution maps, the GAN-generated maps were produced with consistent detail level across the three examples, with high correlation scores (0.891, 0.967 and 0.935) with experimental maps. In contrast, \emph{molmap}-generated maps displayed fine-grained resolution at the level of individual atoms (0.676, 0.619 and 0.653). In practice, these maps effectively captured the spatial distribution of atomic densities but did not seem to distinctly represent SSEs (see on-set panels in Figure \ref{fig:visual_comp}).
In contrast, \emph{struc2mapGAN} seemed to better capture SSEs such as $\alpha$-helices and $\beta$-sheets.
For instance, for EMDB-35136 in Figure \ref{fig:visual_comp}c, the GAN-generated maps distinctly revealed the densities of helical and sheet structures, indicating that our model effectively learned the complexity of experimental maps.
These visual results demonstrate that the maps synthesized by \emph{struc2mapGAN} improved representation of SSEs.

\subsection{Quantitative comparison over resolution}
\begin{figure}[!t]
    \centering
    \includegraphics[width=\linewidth, trim={0cm 0cm 0cm 0cm}, clip]{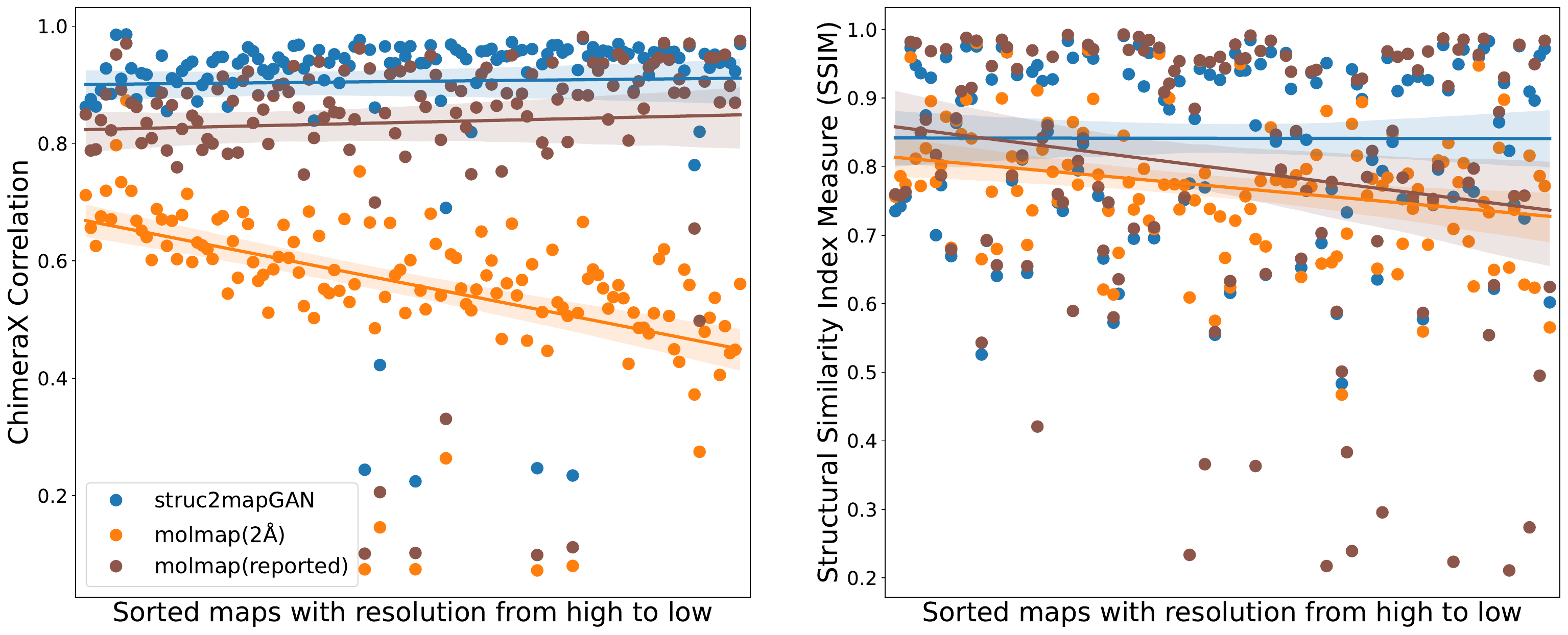}
    \caption{ 
    The scatter plots for comparison of ChimeraX correlation (left) and SSIM (right) for \emph{struc2mapGAN} (blue dots), \emph{molmap} at a resolution cutoff at 2 {\AA} (orange dots), and \emph{molmap} at the reported resolution cutoff (brown dots), across 130 test examples. The test examples were sorted by their reported resolutions from high to low. 
    The shaded area around each colored regression line represents the confidence interval of the regression estimate.
    }
    \label{fig:scatter_plot}
\end{figure}

We then studied how the GAN-generated maps from PDB structures compare against the corresponding experimental counterpart, and how a simulation-based method like \emph{molmap} performs in comparison. For this section, we also aimed to consider the resolution of the experimental map (ranging from high (3 {\AA}) to low (7.9 {\AA})), since GAN-generated maps aim to replicate the map features at high resolution.
As depicted in Figure \ref{fig:scatter_plot}, over 95 \% of GAN-generated maps achieved high correlation scores above 0.8, irrespective of their resolution.
Surprisingly, this correlation was also constant on average across resolution levels (as shown by the fitted curve). A similar result was obtained using SSIM (see Methods).
We hypothesize that the information loss in low-resolution maps, that would result in poor placement of atoms in the PDB structure, was compensated by incorporating high-resolution training data, leading to maintaining performance in low-resolution maps.
To support this hypothesis, we performed the same comparison using \emph{molmap}-simulated maps at both high resolution of 2 {\AA}, and the same reported resolution as the experimental one. In Table \ref{tab:comparison}, we report the mean and median correlation and SSIM scores that indicate better performance of GAN-generated maps. As expected, simulated maps at 2 {\AA} showed poorer correlation and similarity as resolution decreases resulting from less the accurate atomic model (see Figure \ref{fig:scatter_plot}). Simulated maps at reported resolution would maintain constant correlation (compensating atom misplacement by a more diffuse kernel), but with less value than \emph{struc2mapGAN}, and also a decreasing trend for SSIM, supporting that GAN-generated maps can compensate for the uncertainty of atomic positions in PDB structures generated at low resolution.

\begin{table}[ht]
\caption{Performance comparison of different methods.\label{tab:comparison}}
\tabcolsep=0pt
\begin{tabular*}{\textwidth}{@{\extracolsep{\fill}}lcccccccccccccccc@{\extracolsep{\fill}}}
\toprule%
& \multicolumn{2}{@{}c@{}}{SSIM} & \multicolumn{2}{@{}c@{}}{Correlation} & \multicolumn{2}{@{}c@{}}{Correlation about mean} & \multicolumn{2}{@{}c@{}}{PCC} \\
\cline{2-3}\cline{4-5}\cline{6-7}\cline{8-9}%
Methods & Mean & Median & Mean & Median & Mean & Median & Mean & Median \\
\midrule
\textbf{struc2mapGAN}                 & 0.841 & \textbf{0.896} & \textbf{0.906} & \textbf{0.943} & \textbf{0.466} & \textbf{0.502} & \textbf{0.594} & \textbf{0.621} \\
molmap              & 0.771 & 0.774 & 0.559 & 0.573 & 0.315 & 0.311 & 0.452 & 0.470 \\
StructureBlurrer    & 0.764 & 0.771 & 0.603 & 0.620 & 0.335 & 0.334 & 0.475 & 0.499 \\
e2pdb2mrc           & \textbf{0.848} & \textbf{0.896} & 0.613 & 0.649 & 0.322 & 0.325 & 0.483 & 0.519 \\
\bottomrule
\end{tabular*}
\end{table}

\begin{figure*}[ht]
    \centering
    \includegraphics[width=0.9\linewidth, trim={0cm 0cm 0cm 0cm}, clip]{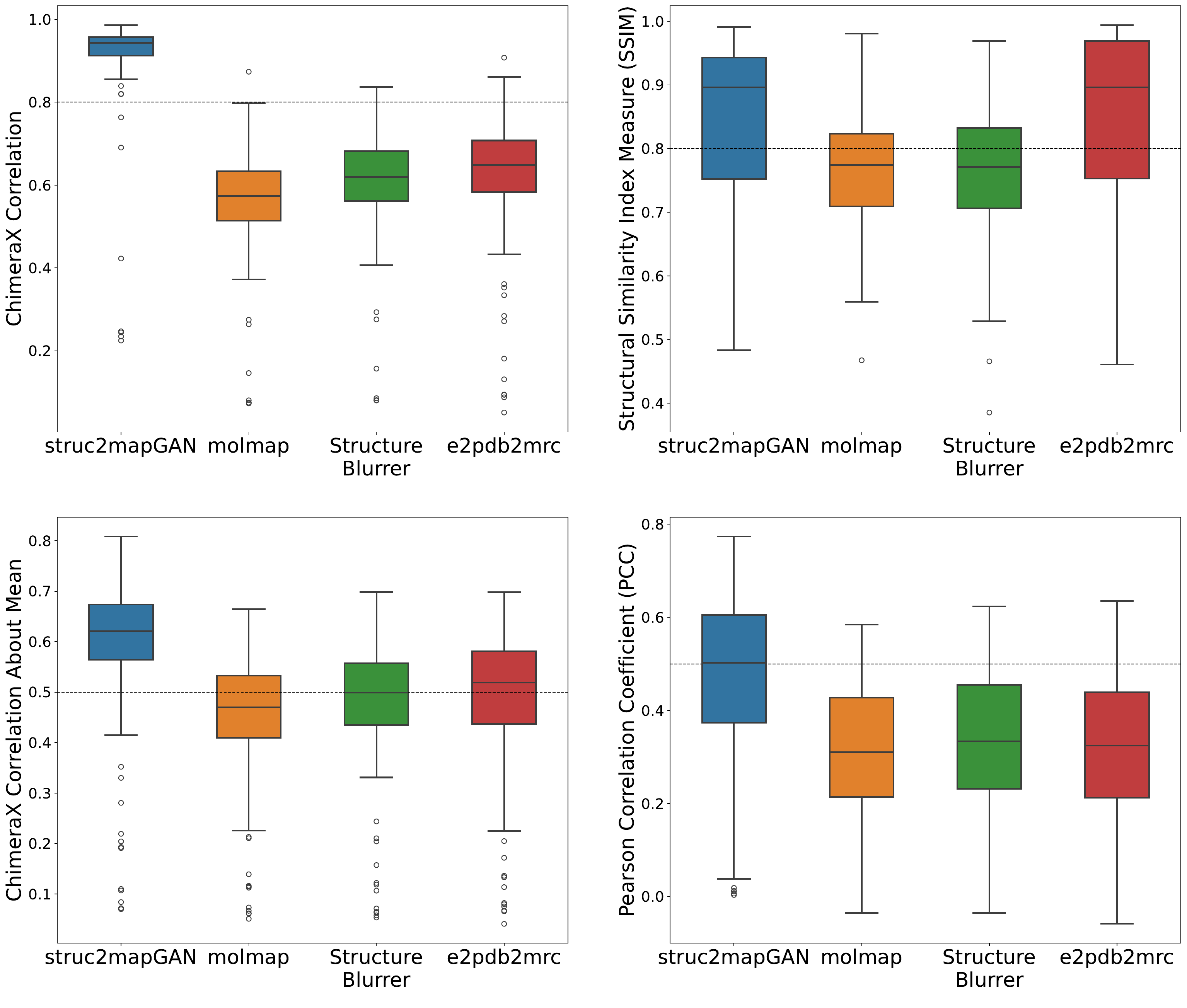}
    \caption{ 
    The box-whisker plots for comparison of different methods (\emph{struc2mapGAN}, \emph{molmap}, \emph{StructureBlurrer}, and \emph{e2pdb2map}) across four evaluation metrics over 130 test examples. For each box-whisker plot, the center line is the median; lower and upper hinges represent the first and third quartiles; the whiskers stretch to 1.5 times the interquartile range from the corresponding hinge; and the outliers are plotted as circles.
    }
    \label{fig:comp_boxplot}
\end{figure*}

\subsection{Benchmarking}

\begin{figure}[ht]
    \centering
    \includegraphics[width=0.75\linewidth, trim={0cm 0cm 0cm 0cm}, clip]{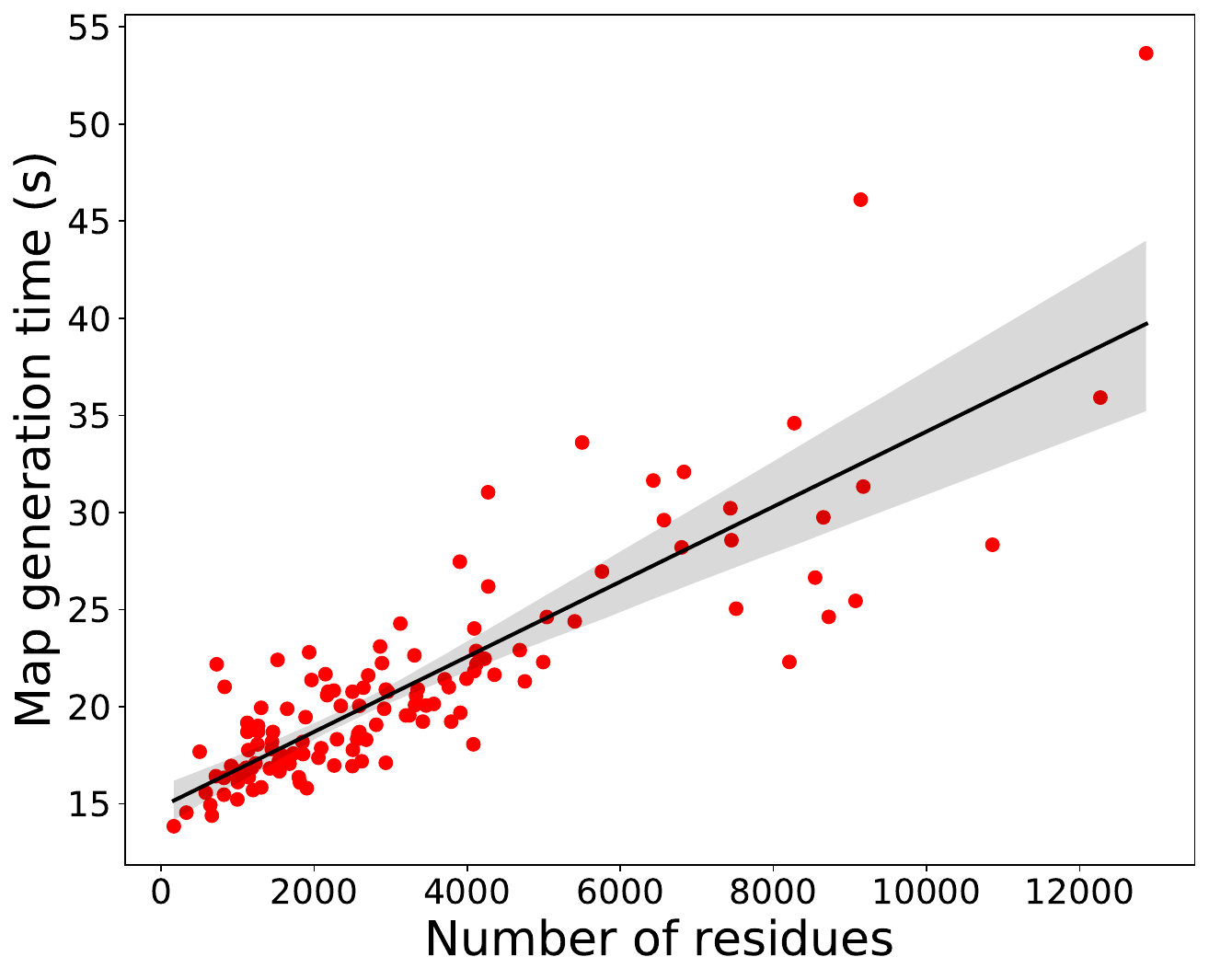}
    \caption{ 
    The scatter plot of map generation time against the number of residues in the molecule. Each red dot represents a molecule's generation based on its residue count. The shaded area around the black regression line represents the confidence interval of the regression estimate. Running times were recorded  for an AMD Ryzen Threadripper 2950X Processor with 32 CPUs.
    }
    \label{fig:inference_time}
\end{figure}

We further compared \emph{struc2mapGAN} with other commonly used simulation-based methods, including \emph{StructureBlurrer} and \emph{e2pdb2mrc}, in terms of SSIM, correlation, and PCC scores across all 130 test examples.
The mean and median values of these metrics are listed in Table \ref{tab:comparison}.
According to the box-and-whisker plots shown in Figure \ref{fig:comp_boxplot}, maps generated by \emph{struc2mapGAN} consistently produced  higher scores in metrics of correlation, correlation about mean, and PCC compared to the other methods. Specifically, \emph{struc2mapGAN} achieved an average correlation of 0.906 and a PCC of 0.594, significantly surpassing the other methods. Although \emph{struc2mapGAN}'s average SSIM score was 0.841, slightly lower than \emph{e2pdb2mrc}'s 0.848, it exhibited a narrower interquartile range, indicating more consistent results. 
The high SSIM scores achieved by \emph{e2pdb2mrc} can be attributed to its effective preservation of local structural details. This preservation is facilitated by the method's requirement to select a larger box size ($500\text{\AA}\times500\text{\AA}\times500\text{\AA}$) than the reference for successfully generating a density map. A larger box size encompasses more spatial context around the molecule, thereby retaining more local structural details. Aside from the the SSIM score for \emph{e2pdb2mrc}, we also observe that all three simulation-based methods yield comparable results across all metrics showing their relative similarity.

\subsection{Map generation time}
To assess the potential use of \emph{struc2mapGAN} in practice, we also recorded the time to generate experimental-like density maps from PDB structures for all 130 test examples. 
Figure \ref{fig:inference_time} shows the wall-clock time plotted against the number of residues of each candidate structure. The scatter plot indicates that the relationship between the map generation time and the number of residues is approximately linear. Upon reviewing several instances, we found that it took around 14 and 20 seconds to generate maps containing 644 and 2501 residues, respectively. For a significantly larger structure with 12868 residues, the map generation time remained within an acceptable range, approximately 53 seconds. These results demonstrate that \emph{struc2mapGAN} scales efficiently with the increasing complexity of protein structures, making it a viable tool for real-time applications.

\section{Ablation study} \label{ablation}
\begin{figure}[!h]
    \centering
    \includegraphics[width=0.8\linewidth, trim={0cm 0cm 0cm 0cm}, clip]{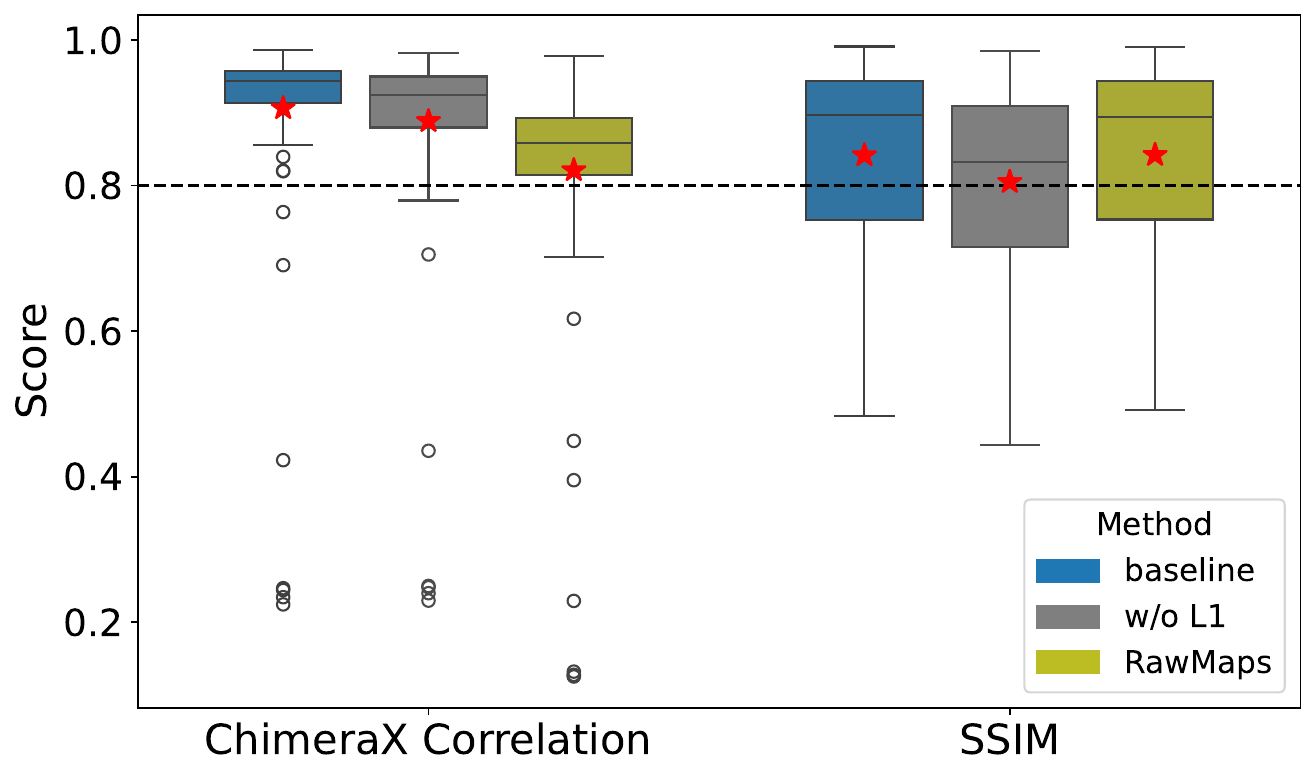}
    \caption{ 
    The box-whisker plots for comparison among \emph{struc2mapGAN} (base), the network trained without \texttt{SmoothL1Loss} (w/o L1), and the network trained with uncurated raw experimental maps (RawMaps). The evaluation was made across two metrics: ChimeraX correlation and SSIM, based on 130 test examples. 
    The red stars represent the mean values of corresponding scores.
    }
    \label{fig:ablation_boxplot}
\end{figure}

Two key factors influence the learning efficiency of \emph{struc2mapGAN}. The first is the incorporation of \texttt{SmoothL1Loss} in the generator as an additional constraint to mitigate mode collapse inherent in GANs and stabilize the training process. The second is the use of curated experimental maps as the learning targets, enabling the model to learn more accurate and reliable mappings between synthetic and experimental maps.
To investigate the impact of these factors on the performance of \emph{struc2mapGAN}, we trained two distinct GAN models: one without \texttt{SmoothL1Loss} (w/o L1) and the other using raw experimental maps without any curation. We then conducted these evaluations on both models using all 130 test examples.
ChimeraX correlation scores are reported in Figure \ref{fig:ablation_boxplot}. The baseline \emph{struc2mapGAN} (i.e. trained with \texttt{SmoothL1Loss} and using curated maps) achieved the highest mean  and median scores, as well as the narrowest interquartile range. These results indicate that incorporating \texttt{SmoothL1Loss} enhances the preservation of overall similarity in the generated maps and leads to more consistent outcomes. 
Similarly, SSIM scores are reported in Figure \ref{fig:ablation_boxplot}, indicating slightly lower mean and median without \texttt{SmoothL1Loss}. 

The model trained with only raw maps achieved SSIM scores with a mean of 0.842 and a median of 0.894, comparable to those of \emph{struc2mapGAN} (mean of 0.841 and  median of 0.896). 
This result was anticipated since the SSIM metric is more sensitive to local structural details. Training with raw maps maintained the local molecular information, resulting in minimal variance in SSIM scores. 
However, most noise, artifacts, and solvents were removed when raw maps were curated, which allows \emph{struc2mapGAN} trained with these curated maps to focus more on overall similarity, thereby yielding higher correlation scores. 
These findings underscore the importance of using an additional \texttt{SmoothL1Loss} to guide network training and employing curated maps as input targets, both of which enhancing the model performance.

\section{Discussion and conclusion}
In this work, we present \emph{struc2mapGAN}, a novel method adapted from the GAN architecture to generate cryo-EM density maps from PDB structures, which mimic the unique characteristics of raw maps.  
We enhance \emph{struc2mapGAN}'s training efficiency by excluding the noise and artifacts and selectively extracting the macromolecule regions from the raw maps as training input, as well as incorporating \texttt{SmoothL1Loss} into the generator.
Our benchmarking results show that \emph{struc2mapGAN} outperforms existing simulation-based methods across various evaluation metrics.
Moreover, its rapid synthesis speed makes it suitable for generating large-scale data.

Overall, by employing \emph{struc2mapGAN}, structural biologists can synthesize improved experimental-like cryo-EM density maps in a timely manner, which can then be used for various applications, such as map-model validation, structure alignment, and guiding the particle-picking process during reconstruction of experimental maps.
Furthermore, machine learning scientists can use \emph{struc2mapGAN} to generate numerous experimental-like density maps as targets, enabling the re-training or fine-tuning of the existing models that were originally trained with simulation-based maps, thereby improving the model performance.

While our results demonstrate superior performance in generating improved density maps, it is not possible to relate these maps to an exact resolution value. In principle, it would be feasible to precisely modulate resolutions by integrating a resolution-conditioned modulation mechanism into the generator. This could involve conditioning the network on a continuous resolution parameter via learnable embedding layers or feature-wise linear modulation (FiLM) \cite{film}, allowing explicit control over output resolution.
Furthermore, with advancements in other generative models \cite{transformer,diffusion}, integrating attention modules into the generator of \emph{struc2mapGAN} could enhance capturing detailed structural information from maps. It would also be interesting to utilize diffusion models for map generation. These will be left for our future work.

Interestingly, our method can also be employed to generate reliable density maps from structures predicted by AlphaFold \cite{alphafold} and other de novo sequence-to-structure generative methods \cite{rosettafold,esmfold}, serving as templates to guide and expedite the particle-picking process during the collection of cryo-EM 2D images for reconstructing experimental density maps \cite{particlepick,recons}. Moreover, these templates have the potential for template matching in cryo-electron tomography \cite{cryoet}. Exploring these applications will also be the focus of our future work.

\section*{Acknowledgments}
This work is supported in parts by a NFRFE-2019-00486 grant.

\section*{Data Availability}
The source code of \emph{struc2mapGAN} is available on GitHub at \url{https://github.com/chenwei-zhang/struc2mapGAN}.
The datasets were derived from sources in the public domain: EMDB databank \cite{EMDB} and PDB databank \cite{PDB}.

\clearpage
\bibliographystyle{unsrt}  
\bibliography{references}

\appendix
\renewcommand{\thetable}{A\arabic{table}}
\setcounter{table}{0} 

\begin{table}[!p]
    \caption{List of all EMDB/PDB examples in training and validation sets. Examples in bold are used as the validation set.}
    \centering
    \includegraphics[width=\linewidth, trim={2cm 5cm 1cm 2cm}, clip]{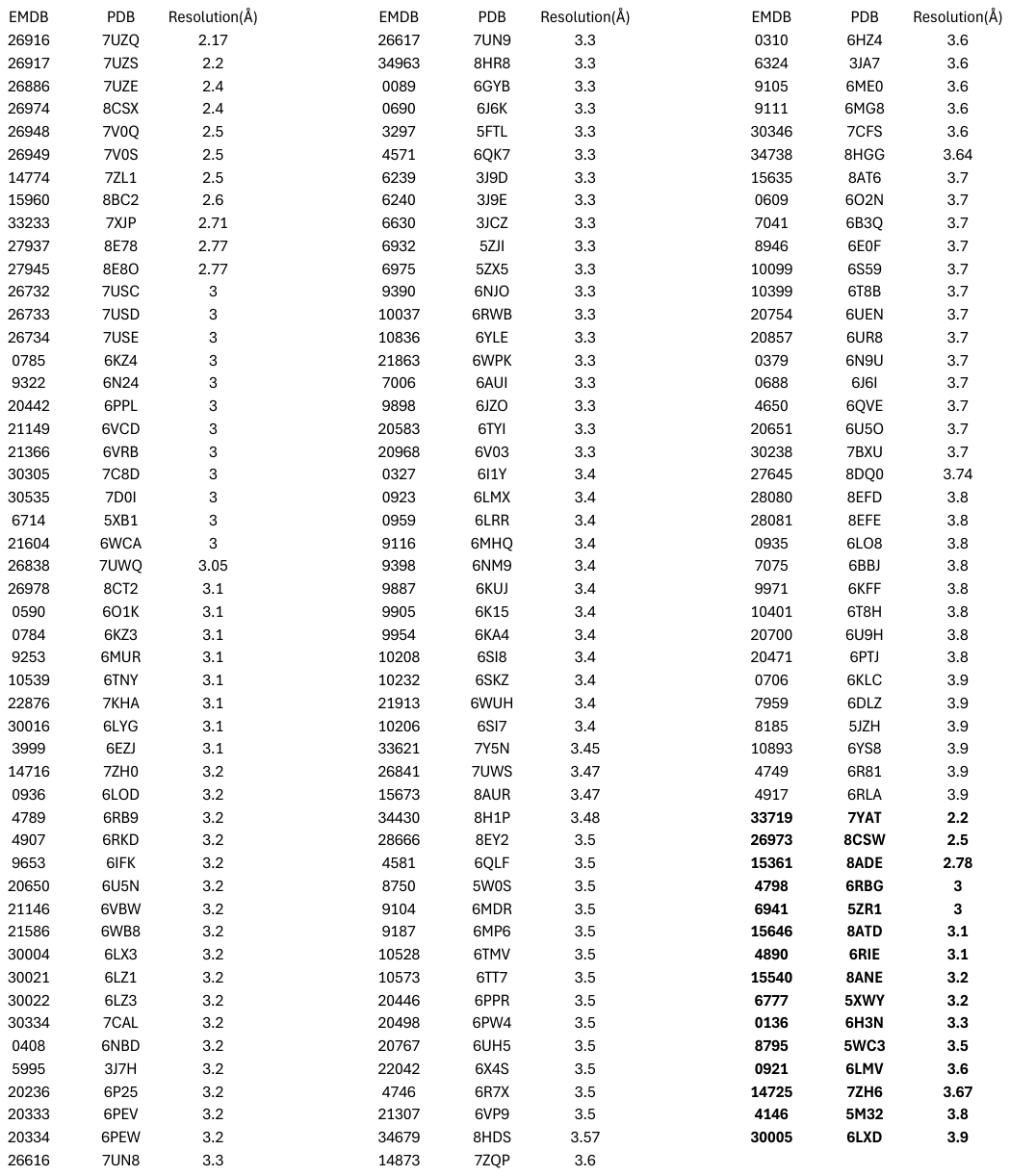}
    \label{tab:trainvaltb}
\end{table}

\begin{table}[!p]
    \caption{List of all EMDB/PDB examples in the test set.}
    \centering
    \includegraphics[width=\linewidth, trim={2cm 5cm 1cm 2cm}, clip]{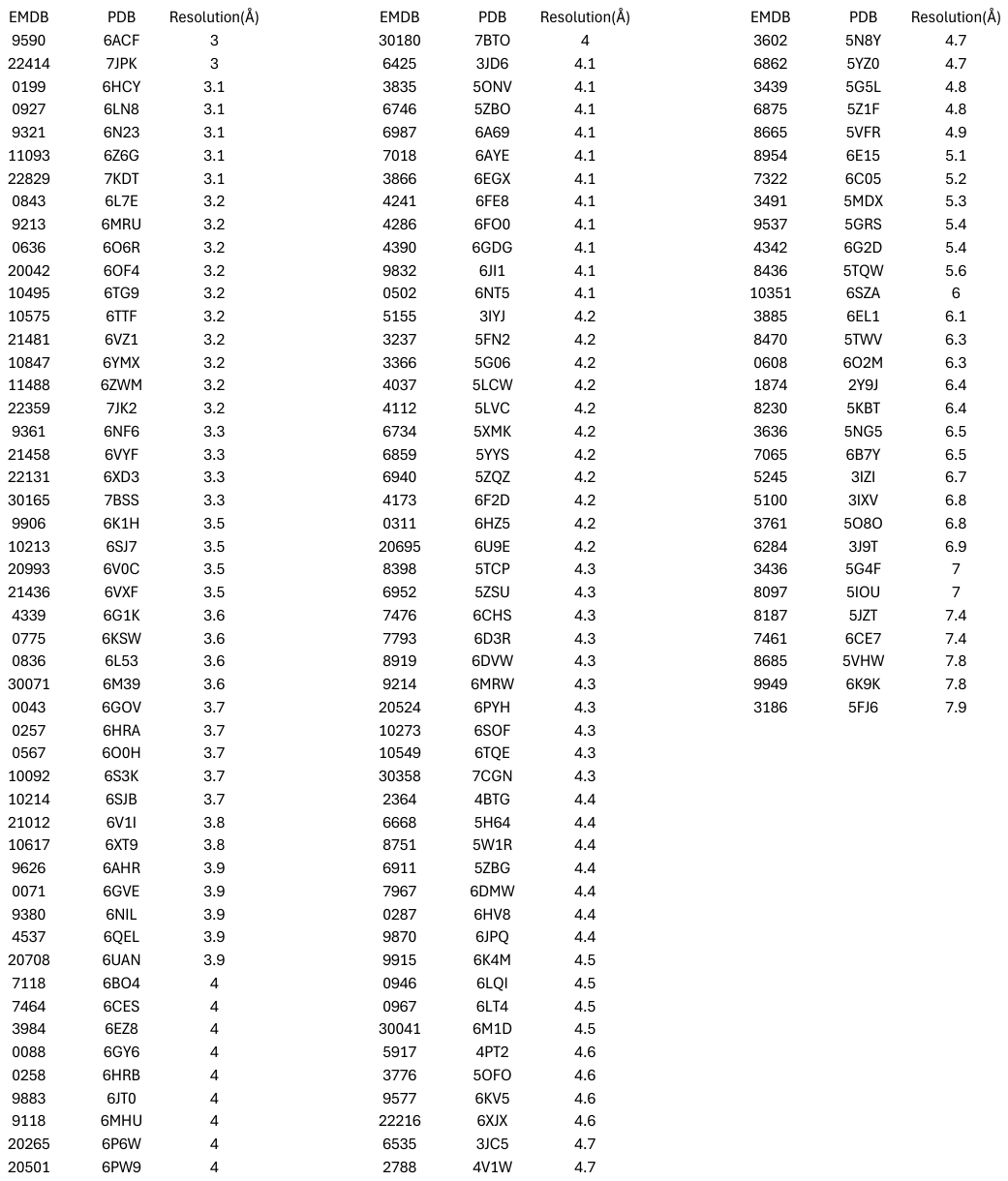}
    \label{tab:testtb}
\end{table}

\end{document}